\documentclass[11pt]{article}

\usepackage[final]{acl}
\usepackage{amsmath}
\usepackage{booktabs}
\usepackage{multirow}
\usepackage{amssymb}
\usepackage{array}
\usepackage[table]{xcolor}
\usepackage{arydshln}
\usepackage[utf8]{inputenc}
\usepackage{xcolor}
\usepackage{colortbl}
\usepackage{nicematrix}
\usepackage{tcolorbox}
\usepackage{listings}

\definecolor{rowgray}{gray}{0.95}      
\definecolor{rowblue}{HTML}{F2F9FF}    
\definecolor{gaincolor}{RGB}{55,126,184}
\definecolor{wincolor}{HTML}{B03A2E}

\usepackage{times}
\usepackage{latexsym}

\usepackage[T1]{fontenc}

\usepackage[utf8]{inputenc}

\usepackage{microtype}

\usepackage{inconsolata}

\usepackage{graphicx}

\title{\textsc{Revealer}: Reinforcement-Guided Visual Reasoning for Element-Level Text-Image Alignment Evaluation}

\author{
 \textbf{Fulin Shi\textsuperscript{1}},
 \textbf{Wenyi Xiao\textsuperscript{1}},
 \textbf{Bin Chen\textsuperscript{2}},
 \textbf{Liang Ding\textsuperscript{2}},
 \textbf{Leilei Gan\textsuperscript{1}\thanks{Corresponding author}}
\\
 \textsuperscript{1}Zhejiang University,
 \textsuperscript{2}Alibaba Group
 \\
\texttt{\{fulinshi, leileigan\}@zju.edu.cn}
}

\begin{document}
\maketitle
\begin{abstract}
Evaluating the alignment between textual prompts and generated images is critical for ensuring the reliability and usability of text-to-image (T2I) models. However, most existing evaluation methods rely on coarse-grained metrics or static Question Answering (QA) pipelines, which lack fine-grained interpretability and struggle to reflect human preferences. To address this, we propose \textbf{\textsc{Revealer}}, a reinforcement-guided visual reasoning framework for element-level text-to-image alignment evaluation. 
Adopting a structured ``grounding–reasoning–conclusion'' paradigm, our method enables Multimodal Large Language Models (MLLMs) to explicitly localize semantic elements and derive interpretable alignment judgments. We optimize the model via Group Relative Policy Optimization (GRPO) using a multi-dimensional reward function that targets format compliance, localization precision, and alignment accuracy.
Extensive experiments confirm that \textsc{Revealer} achieves state-of-the-art results across four benchmarks. Notably, on EvalMuse-40K, it surpasses the strong proprietary Gemini 3 Pro and Training-based baselines with absolute accuracy gains of \textbf{+4.0\%} and \textbf{+13.1\%}, respectively. 
Ablation studies further demonstrate the efficacy of our method, contributing a cumulative \textbf{19.4\%} improvement over the base model. 
\href{https://anonymous.4open.science/r/F698/}{Code: https://anonymous.4open.science/r/F698/}
\end{abstract}

\section{Introduction}

\begin{figure}[t]
  \includegraphics[width=0.47\textwidth]{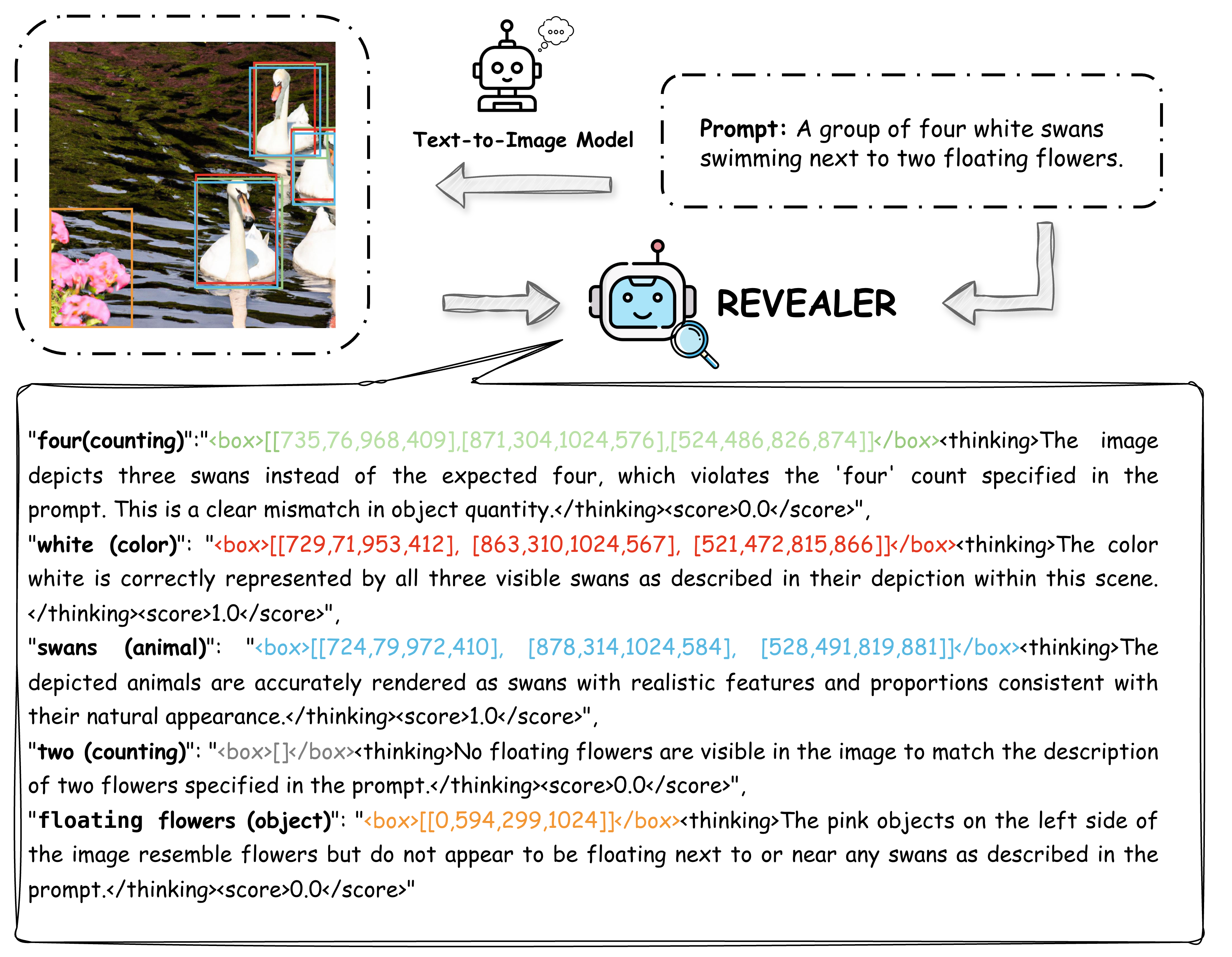}
  \caption{\textbf{\textsc{Revealer}} performs element-level text-to-image alignment evaluation via structured visual reasoning, following a grounding–reasoning–conclusion paradigm.}
  \label{fig:case}
\end{figure}
Text-to-image (T2I) models \cite{sdxl} such as DALL·E \cite{dall}, Stable Diffusion \cite{sd, sd3}, and Imagen \cite{imagen} have made significant strides in generating visually appealing and semantically rich images from natural language prompts. 
With the widespread adoption of T2I models, ensuring that the generated image faithfully aligns with the semantics of the input text becomes increasingly critical, which is known as the task of \textit{text-image alignment evaluation}.

Early text-image alignment evaluation methods ~\cite{fid, clipscore, v3} rely on coarse-grained metrics that collapse rich semantic structures into single scalar scores (e.g., CLIPScore ~\cite{clipscore}), but they lack interpretability and are often insensitive to fine-grained mismatches, such as object count, attributes, and spatial composition.
To improve the interpretability of CLIPScore, VIEScore ~\cite{vie} proposes leveraging multimodal large language models (MLLMs) ~\cite{llava, qwenvl, instructblip} to generate natural language rationales alongside alignment scores.
To facilitate fine-grained alignment evaluation, question-answering (QA)-based approaches ~\cite{t2icombench, llmscore}, such as TIFA~\cite{tifa} and $\text{VQ}^2$~\cite{vq2}, employ off-the-shelf large language models (LLMs) to generate multiple verifiable questions from the input prompts, with each question targeting a distinct facet of alignment evaluation.
However, due to their reliance on predefined question templates, these methods often fail to generate questions that adequately assess the alignment of all elements in the prompt, especially in complex cases.
Moreover, most MLLM-based QA alignment evaluation methods rely solely on prompt engineering without dedicated supervision, resulting in suboptimal evaluation performance.
To address these issues, EvalMuse-40K~\cite{evalmuse} introduces a large-scale benchmark featuring element-level binary annotations (e.g., objects, attributes, locations), providing rich supervision for fine-grained alignment training and evaluation.
However, its annotations are often treated as classification tasks, lacking interpretable reasoning paths. 
Recently, UnifiedReward-R1~\cite{umcotrm} has explored reinforcement learning to enable chain-of-thought-style text-image alignment score prediction by incorporating rule-based reward signals.
However, UnifiedReward-R1 only provides an overall alignment score for each evaluated dimension, and lacks the capability to explicitly determine whether specific objects or elements are correctly generated according to the input prompt.

To address the aforementioned limitations, we propose \textbf{\textsc{Revealer}}, a reinforcement-guided visual reasoning framework for element-level text-to-image alignment evaluation.
As illustrated in Figure~\ref{fig:case}, \textsc{Revealer} operates through a three-stage framework comprising grounding, reasoning, and conclusion, which emulates human-like analysis in text-image alignment evaluation.
At the first stage, the visual reasoning grounds each element of the prompt to specific regions within the generated images, thereby providing essential contextual information for alignment reasoning. 
Here, the elements are derived by decomposing the input prompt into fine-grained semantic units, which follows the TIFA taxonomy categorization (e.g., object, attribute, activity, etc.).
At the second stage, a free-form natural language explanation is produced to evaluate the alignment between the grounded visual content and the corresponding element in the prompt.
Finally, an element-level alignment score is derived by comprehending the information obtained from the grounding and reasoning stages.
This interleaved visual-textual reasoning process significantly improves the interpretability of the evaluation metric, while simultaneously offering dense supervision signals for model training.

To equip the MLLM with the capability to follow the three-stage visual reasoning paradigm, we first fine-tune it on automatically curated visual reasoning trajectories. 
Subsequently, a reinforcement learning (RL) phase—implemented via Group Relative Policy Optimization (GRPO)~\cite{grpo}—is employed to bolster the model's reasoning capabilities. 
Specifically, we design a comprehensive rule-based reward function to leverage the rich supervision signals intrinsic to all three stages. 
To facilitate this training recipe, we propose an automated data curation pipeline that synthesizes training trajectories by synergizing an expert vision model with general-purpose LLMs.

Extensive experiments across four benchmarks demonstrate that \textsc{Revealer} achieves state-of-the-art performance. Specifically, our method yields substantial accuracy gains relative to the Training-based baseline, achieving increases of \textbf{+13.1\%} on EvalMuse-40K, \textbf{+9.4\%} on RichHF, \textbf{+6.3\%} on MHaluBench, and \textbf{+6.5\%} on GenAI-Bench. Notably, it surpasses the strong proprietary model, Gemini 3 Pro, by a margin of \textbf{+4.0\%} on EvalMuse-40K. 
Ablation studies further validate the efficacy of our framework components, showing a cumulative performance boost of \textbf{+19.4\%} over the base model, while subsequent analyses confirm that explicit visual reasoning enhances both fine-grained alignment accuracy and interpretability.

\section{Related Work}
This section provides a brief review of related work.

\noindent\textbf{Automated Methods and Metrics for Text-Image Alignment Evaluation.} Early metrics such as \cite{clipscore, blip2, pickapic} evaluate text-image alignment via cosine similarity in embedding space. While computationally efficient, these approaches lack sensitivity to fine-grained mismatches.
To improve interpretability, structured evaluation methods such as TIFA \cite{tifa} and $\text{VQ}^2$ \cite{vq2} convert prompts into QA or NLI tasks, though their performance depends heavily on hand-crafted templates. More recent efforts introduce stronger compositional reasoning: VIEScore \cite{vie} uses instruction-following MLLMs to generate alignment scores with natural language rationales; DSG \cite{dsg} leverages semantic scene graphs for robustness; And VQAScore \cite{vqascore} decomposes prompts into atomic QA sub-tasks for modular evaluation.
FGA-BLIP2 \cite{evalmuse} fine-tunes models for element-level alignment scoring, while PN-VQA \cite{evalmuse} adopts a prompt-based querying strategy without fine-tuning. A recent task-decomposed framework \cite{t2ieval} further enhances interpretability and robustness by combining modular pipelines with multi-perspective metrics.
In parallel, MLLM-based methods \cite{supervisedfinetuningmultimodal} directly predict alignment scores through supervised finetuning on human-aligned data.

\noindent\textbf{Reinforcement Learning for Visual Reasoning and Evaluation.} Reinforcement learning (RL) has been used to enhance alignment evaluation, as in T2I-Eval-R1~\cite{t2ievalr1}, UM-CoT-RM~\cite{umcotrm}, Unified Hallucination Detection~\cite{unifiedhallucination}, UnifiedReward~\cite{wang2025unifiedrewardmodel} and Vision-R1~\cite{visionr1}, which aim to enhance alignment consistency in visual content and improving interpretability through reasoning chains.
RL also improves visual reasoning: DeepEyes~\cite{deepeyes} and OpenThinkIMG~\cite{openthink} train agents for spatial reasoning, and Q-Insight~\cite{qinsight} applies reinforcement learning to train visual agents for interpretable image quality assessment. ViLaSR~\cite{vilasr} reinforces geometric understanding, and works like Thinking with Generated Images~\cite{thinkinggeneratedimages}, Chain-of-Focus~\cite{cof}, and UniVG-R1~\cite{univgr1} explore internal reasoning via sketching, zooming, or CoT-based image generation. These efforts demonstrate how sequential visual reasoning enhances robustness and explainability.

\begin{figure*}[t]
  \centering
  \includegraphics[width=1.0\textwidth]{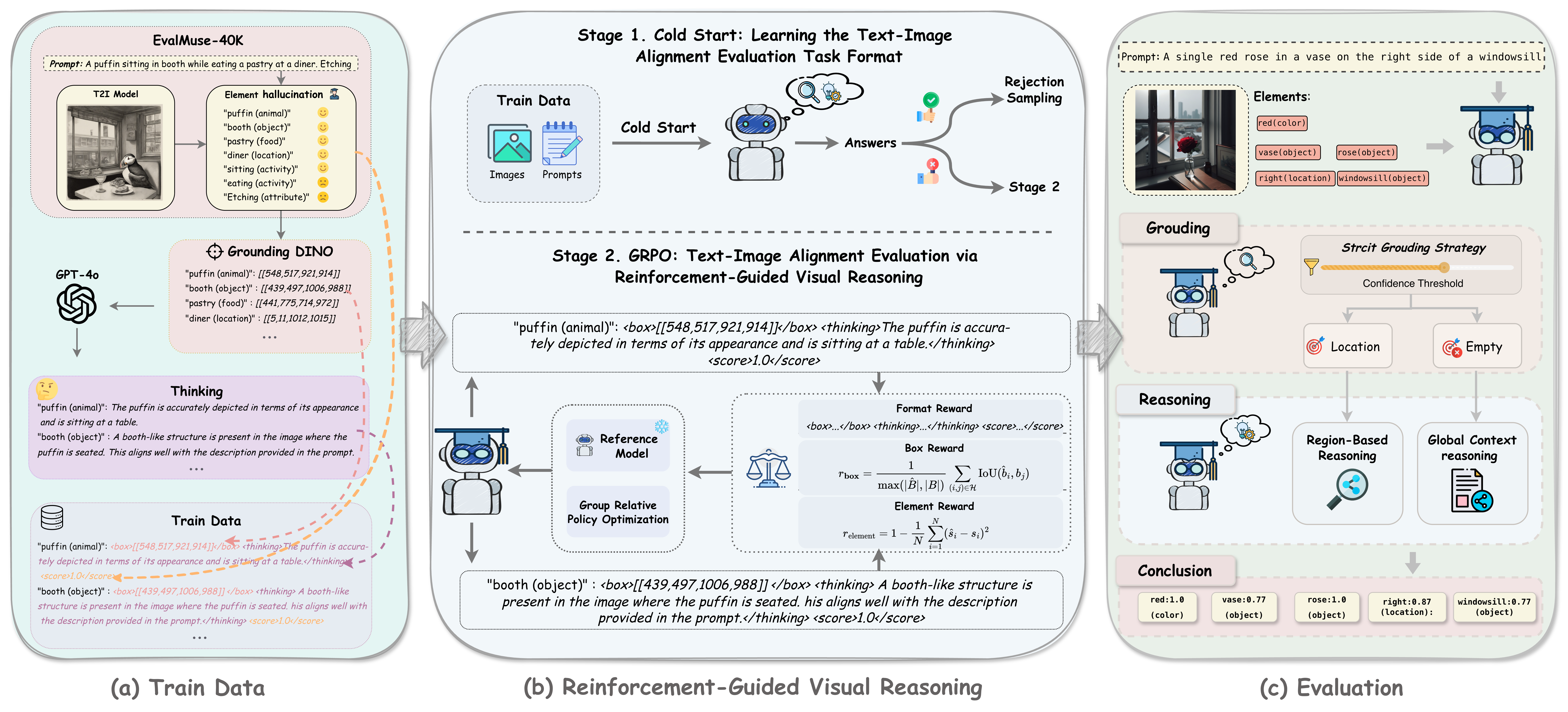}
  \caption{Our work consists of three components: (a) Training data is constructed using Grounding DINO and GPT-4o to generate structured alignment annotations; (b) A two-stage training pipeline performs reinforcement-guided visual reasoning via GRPO; (c) The model is evaluated on four fine-grained alignment benchmarks.}
  \label{fig:bigframework}
\end{figure*}

\section{Methodology}
In this section, we first introduce the visual reasoning process for element-level text-image alignment evaluation $\S \ref{sec:visual_reasoning}$. 
We then describe the training dataset curation procedure $\S \ref{sec:sft}$, followed by a detailed illustration of the two-stage training methodology $\S \ref{sec:rl}$. 
The overall methodology is illustrated in Figure~\ref{fig:bigframework}.
\subsection{Visual Reasoning for Element-Level Text-Image Alignment Evaluation}
\label{sec:visual_reasoning}
Despite recent advances \cite{deepeyes, openthink, qinsight}, existing approaches to T2I alignment evaluation still struggle with accurately assessing element-level alignment between textual descriptions and generated images.
Inspired by the human-like alignment analysis process, which follows a three-stage chain-of-thought ``grounding---reasoning---conclusion'', we propose visual reasoning guided element-level text-image alignment evaluation via reinforcement learning.

Specifically, the visual reasoning process unfolds in three stages, each corresponding to a structured component in the reasoning trajectory. 
In the \textbf{grounding} stage, the sequence begins with a special token \texttt{<box>}, followed by a predicted bounding box list that localizes a semantic element from the input prompt within the generated image. 
Next, in the \textbf{reasoning} stage, the \texttt{<thinking>} token precedes a free-form natural language explanation that evaluates the semantic alignment between the visual content in the localized region and the corresponding element in the prompt. 
Finally, in the \textbf{conclusion} stage, the sequence begins with the \texttt{<score>} token followed by a scalar alignment score $s \in [0, 1]$, 
where the scalar magnitude quantifies the degree of visual-semantic consistency, with higher values signifying superior alignment.

This three-stage visual reasoning alignment evaluation offers several notable advantages. 
First, by explicitly localizing specific semantic elements within the generated image, the grounding stage facilitates more precise visual-textual alignment and provides essential contextual information for subsequent reasoning.
Second, the intermediate natural language rationales generated in the the reasoning stage enhance the interpretability of the final alignment score.
Lastly, this staged visual reasoning yields rich supervision signals for both training and evaluation, as will be further detailed in the following sections.

\noindent\textbf{Visual Reasoning Trajectory Curation.} To support the aforementioned visual reasoning training, we propose an automated method for curating such visual reasoning trajectory, which combines the visual grounding capability of an expert model and the reasoning ability of proprietary LLMs. 
The overall curation process is shown in Figure \ref{fig:bigframework} (a).

The visual reasoning dataset is derived from the training split of EvalMuse-40K. 
EvalMuse-40K is a large-scale benchmark for text-to-image alignment evaluation, which contains 40K image–prompt pairs with element-level binary annotations. 

Specifically, let $(\mathcal{I}, \mathcal{P}, \{\langle e_i, a_i \rangle\}_{i=1}^N)$ denote a data point in EvalMuse-40K, where $\mathcal{I}$ is the generated image and $\mathcal{P}$ is the input prompt. 
$\{\langle e_i, a_i \rangle\}_{i=1}^N$ corresponds to the set of the element-level annotations (e.g., objects, attributes, locations), where $e_i$ denotes an element, and $a_i$ denotes the binary answer. 
The visual reasoning trajectory for each data point is constructed as follows. 
First, for each $e_i$ in the set ${\langle e_i, a_i \rangle}_{i=1}^N$, we utilize Grounding DINO~\cite{groundingdino}, a state-of-the-art object grounding model, to associate the element with a corresponding region in the generated image $\mathcal{I}$. 
This grounding step produces a list of bounding boxes $\{b_{i,j} = [x_1, y_1, x_2, y_2]\}_{j=1}^{K_i}$ that spatially localize the element $e_i$ within the image, where $K_i$ denotes the number of detected regions associated with $e_i$. 
We employ a \textit{strict grounding strategy} by raising the detection confidence threshold, which yields an empty list for low-confidence regions, effectively preventing error propagation caused by incorrect localization (see Sec.~\ref{sec:strict strategy} for details). 
A detailed analysis of the bounding box annotation quality is presented in Sec.~\ref{sec:validation of Visual Grounding}.

Subsequently, for each element $e_i$, GPT-4o is conditioned on the input tuple $(\mathcal{I}, \mathcal{P}, e_i, b_i)$ to generate a natural language explanation $r_i$ and a predicted alignment label $\hat{a}_i$. 
If the associated bounding box set $b_i$ is an empty list (\texttt{[]}), the model is explicitly prompted to perform reasoning based on the global visual context of the image $\mathcal{I}$. 
Otherwise, the reasoning focuses on the specific localized regions.
To ensure the high quality of the generated reasoning rationales $r_i$, we employ a two-stage quality assurance strategy to strictly filter out low-quality samples (see Appendix~\ref{app:quality_assurance} for details).
By following the above procedures, we finally curate a dataset comprising 25K high-quality samples, each annotated with a three-stage visual reasoning trajectory, denoted as $\mathcal{D}_\text{VisualReason}$.

\subsection{Cold-Start Training with Automatically Constructed Visual Reasoning Trajectory.}
\label{sec:sft}
To enable the MLLM to follow the proposed three-stage visual reasoning format, we first introduce a cold start training phase, in which the MLLM is fine-tuned on the automatically constructed visual reasoning trajectory $\mathcal{D}_\text{VisualReason}$.

Specifically, we sample a subset of 5,000 annotated instances from $\mathcal{D}_\text{VisualReason}$, comprising 2,500 real and 2,500 synthetic image–prompt pairs, denoted as $\mathcal{D}_\text{SFT}$.
The selected samples are curated to ensure diversity across a wide range of element types, such as objects, attributes, and spatial. 
The cold start training uses supervised fine-tuning (SFT) to minimize the negative log-likelihood (NLL) of the token sequence.
Formally, given $\mathcal{D}_\text{SFT}$, the model is trained on it to output a structured sequence of the form:
$<$$\boldsymbol{box}$$>$[[x$_1$, y$_1$, x$_2$, y$_2$], ...]$<$$/\boldsymbol{box}$$>$
$<$$\boldsymbol{thinking}$$>$reasoning process$<$$/\boldsymbol{thinking}$$>$
$<$$\boldsymbol{score}$$>$$s\in[0, 1]$$<$$/\boldsymbol{score}$$>$,
where $<$$\boldsymbol{box}$$>$ denotes the predicted bounding box, $<$$\boldsymbol{thinking}$$>$ is a free-form explanation, and $<$$\boldsymbol{score}$$>$ reflects the degree of alignment, with lower scores indicating stronger misalignment.
The objective is to minimize the standard negative log-likelihood (NLL) loss of the structured reasoning sequence conditioned on the input image and prompt. The detailed mathematical formulation is provided in Appendix~\ref{sec:training_objectives}.

This cold start training phase equips the model with the ability to follow the visual reasoning format, establishing a baseline for subsequent RL-based optimization.

\subsection{Visual Reasoning for Element-Level Text-Image Alignment Evaluation via Reinforcement Learning}
\label{sec:rl}
While cold start training provides a baseline for element-level text-image alignment, its ability to incentivize deep reasoning capabilities in foundation models has been shown to be inferior to that of reinforcement learning \cite{t2ievalr1, umcotrm, visionr1}.
To further enhance the model's visual reasoning performance, we introduce an RL stage based on GRPO~\cite{grpo}, equipped with a task-specific reward function and a challenging-sample selection strategy.

\noindent\textbf{Challenging-sample Selection.}
To improve training quality, we retain only challenging samples for the reinforcement learning stage. 
Specifically, we use the cold-start model to generate alignment predictions on $\mathcal{D}_\text{VisualReason}$, and filter out data where the model accurately judges the alignment status of all elements. 
Only examples with at least one incorrectly predicted element are retained for the GRPO training. 
This results in a curated subset of 20K hard cases from the EvalMuse-40K dataset, denoted as $\mathcal{D}_\text{Challenging-Sample}$, used to optimize the model’s alignment policy.

\noindent\textbf{Reward Shaping.} 
Given the rich supervision signals in $\mathcal{D}_\text{Challenging-Sample}$, we construct a composite reward function to guide the model’s behavior along multiple dimensions:

\noindent\textbf{(1) Format Reward} evaluates whether the generated output adheres to the required structured format, including grounding stage ($<$$\boldsymbol{box}$$>$$<$$/\boldsymbol{box}$$>$), reasoning stage ($<$$\boldsymbol{thinking}$$>$$<$$/\boldsymbol{thinking}$$>$), and conclusion stage ($<$$\boldsymbol{score}$$>$$<$$/\boldsymbol{score}$$>$). 
  Specifically, we assign a binary reward $r_{\text{format}} \in \{0,1\}$, where $r_{\text{format}} = 1$ if the output format is correct and $r_{\text{format}} = 0$ otherwise.

\noindent\textbf{(2) Box Reward} quantifies the localization accuracy of predicted bounding boxes by comparing them with ground-truth annotations. 
It adopts a commonly used matching-based strategy to compute the Intersection over Union (IoU) between predicted and ground-truth boxes.
Specifically, let $\hat{B}$ and $B$ be the predicted and ground-truth bounding box sets for each element. 
We first compute the pairwise IoU matrix $\mathbf{M}$, then apply the Hungarian Algorithm to find the optimal one-to-one match as follows:
\begin{equation}
r_{\text{box}} = \frac{1}{\max(|\hat{B}|, |B|)} \sum_{(i,j) \in \mathcal{H}} \text{IoU}(\hat{b}_i, b_j)
\end{equation}
where $\mathcal{H}$ is the set of matched pairs returned by the Hungarian algorithm, and unmatched elements are assigned zero IoU.

\noindent\textbf{(3) Element Reward} evaluates the fine-grained accuracy of the predicted alignment scores. 
Instead of using a simple absolute difference, we adopt a squared-error based formulation to impose heavier penalties on large deviations.
Specifically, for each element, the predicted scalar score $\hat{s}_i$ is compared against the reference score $s_i$ as follows:
\begin{equation}
r_{\text{element}} = 1 - \frac{1}{N} \sum_{i=1}^{N} (\hat{s}_i - s_i)^2
\end{equation}
This formulation yields a continuous reward in the range $[0, 1]$. By utilizing the squared term, the reward provides sharper gradients for significant errors, encouraging the model to converge more strictly toward the ground truth compared to linear feedback.

By combining the aforementioned rewards, the total reward for RL training is defined as:
\begin{equation}
r(\tau) = \lambda_1 r_{\text{format}} + \lambda_2 r_{\text{box}} + \lambda_3 r_{\text{element}}
\end{equation}
where $\lambda_1$, $\lambda_2$, and $\lambda_3$ are weighting hyperparameters tuned via grid search (see Appendix~\ref{sec:training_objectives} for details).

\noindent\textbf{Reinforcement Optimization.}
For policy optimization, GRPO samples a group of outputs $\{o_i\}_{i=1}^G$ for each query $q$ and utilizes group-based advantage normalization. The policy $\pi_\theta$ is updated by maximizing the following surrogate objective:

\begin{equation}
\footnotesize
\small
\begin{split}
\mathcal{J}_{GRPO}(\theta) &= \mathbb{E}{[q \sim \mathcal{D}_\text{Challenging-Sample}, \{o_i\}_{i=1}^G \sim \pi_{\theta_{old}}(O|q)]} \\
&\quad \frac{1}{G}\sum_{i=1}^G\frac{1}{|o_i|} \sum_{t=1}^{|o_i|} \left\{ \min \left[ \rho_t A_t, \text{clip}(\cdot) \cdot A_t \right] \right. \\
&\quad \left. - \beta \mathbb{D}_{\text{KL}}\left[\pi_{\theta} || \pi_{\text{ref}}\right]\right\} ,
\end{split}
\label{eq:GRPO-obj}
\end{equation}

\noindent where $\rho_t$ denotes the policy ratio $\frac{\pi_\theta(o_t \mid q)}{\pi_{\theta_{\text{old}}}(o_t \mid q)}$, $A_t$ represents the advantage normalized within the group, and $\mathbb{D}_{\text{KL}}$ is the unbiased KL-divergence estimator \citep{kl_approx}. $\text{clip}(\cdot)$ refers to applying a clipping function to $\rho_t$ that bounds it within $[1 - \epsilon, 1 + \epsilon]$, where $\epsilon$ is hyperparameter. Full mathematical derivations and implementation details are provided in Appendix~\ref{sec:training_objectives}.

\section{Experiments}
\subsection{Experimental Setup}
\noindent\textbf{Evaluation Benchmarks.} We conduct experiments on four fine-grained benchmarks: EvalMuse-40K, RichHF, MHaluBench, and GenAI-Bench. Details are included in~\ref{sec: benchmark}.

\noindent\textbf{Evaluation Metrics.} To comprehensively assess model performance, we employ three metrics across all benchmarks: Spearman's Rank Correlation Coefficient (SRCC) and Pearson Linear Correlation Coefficient (PLCC) to measure the correlation between predicted scores and human judgments, alongside Accuracy (ACC) for binary classification evaluation.

\noindent\textbf{Baselines.}
We compare our method with a series of strong baselines from two categories:
\textbf{(1) Prompting-based Methods.} We include four representative approaches for text-to-image alignment evaluation: TIFA~\cite{tifa}, $\text{VQ}^2$~\cite{vq2}, VIEScore~\cite{vie}, and VQAScore~\cite{vqascore}. 
Additionally, we introduce a training-free variant of \textbf{\textsc{Revealer}}, where Grounding DINO is utilized to extract object regions, which are subsequently passed to Gemini 3 Pro for reasoning and scalar alignment scoring.
\textbf{(2) Training-based Methods.} 
We include FGA-BLIP2~\cite{evalmuse}, a specialized end-to-end scoring model fine-tuned on the EvalMuse-40K training set to directly predict element-level alignment scores.
Additionally, we establish strong supervised baselines using general MLLMs: Qwen3-VL-8B-Instruct, InternVL3-8B-hf, and LLaVA-v1.6-Mistral-7B-hf. 
These models are fully fine-tuned on $\mathcal{D}_{\text{VisualReason}}$ to generate the complete visual reasoning trajectory (grounding, reasoning, and conclusion).

\noindent\textbf{Implementation Details.}
All models are trained using 8$\times$NVIDIA H200 GPUs. 
Our framework demonstrates exceptional training efficiency with low resource consumption: the SFT stage (5 epochs) requires approximately 16 GPU hours, while the RL stage (3 epochs) consumes around 120 GPU hours.

\setlength{\tabcolsep}{0.35mm}
\begin{table*}[ht]
\centering
\small
\renewcommand{\arraystretch}{1.1} 
\begin{tabular}{
  ll 
  cc c<{\hspace{6pt}}        
  >{\hspace{6pt}}c cc<{\hspace{6pt}} 
  >{\hspace{6pt}}c cc<{\hspace{6pt}} 
  >{\hspace{6pt}}c cc       
}
\toprule
\noalign{\vskip 2pt}
\multirow{2}{*}{\textbf{Method}} & \multirow{2}{*}{\textbf{Model}} &  
\multicolumn{3}{c<{\hspace{6pt}}}{\textbf{EvalMuse-40K}} & 
\multicolumn{3}{>{\hspace{6pt}}c<{\hspace{6pt}}}{\textbf{RichHF}}   & 
\multicolumn{3}{>{\hspace{6pt}}c<{\hspace{6pt}}}{\textbf{MHaluBench}} & 
\multicolumn{3}{>{\hspace{6pt}}c}{\textbf{GenAI-Bench}}\\
\cmidrule(lr){3-5} \cmidrule(lr){6-8} \cmidrule(lr){9-11} \cmidrule(lr){12-14}
& & \textbf{srcc} & \textbf{plcc} & \textbf{acc} & \textbf{srcc} & \textbf{plcc} & \textbf{acc} & \textbf{srcc} & \textbf{plcc} & \textbf{acc} & \textbf{srcc} & \textbf{plcc} & \textbf{acc}\\
\noalign{\vskip 2pt}
\hline
\multicolumn{14}{c}{\textit{\textbf{Prompting-based Methods}}} \\
\hline
\multirow{3}{*}{TIFA} & Gemini 3 Pro & 68.1 & 65.8 & 81.3  & 66.1 & 65.4 & 80.8 & 68.5 & 67.2 & 81.0 & 71.4 & 72.3 & 83.9\\
& Qwen3-VL-235B-A22B-Instruct  & 66.3 & 65.1 & 80.4  & 64.6 & 63.9 & 80.5 & 67.8 & 66.8 & 81.7 & 68.4 & 71.5 & 83.2\\
\cline{2-14}
\multirow{3}{*}{$\text{VQ}^2$}& GPT-4o  & 67.9 & 66.4 & 81.7  & 63.9 & 64.8 & 77.9 & 65.8 & 67.4 & 80.7 & 70.2 & 71.4 & 84.1\\
& Qwen3-VL-235B-A22B-Instruct  & 68.1 & 66.9 & 80.9  & 64.4 & 63.1 & 80.3 & 67.2 & 65.6 & 81.5 & 70.7 & 71.8 & 83.0\\
\cline{2-14}
VQAScore & CLIP-FlanT5-XXL & 51.8 & 51.2 & 65.5  & 63.9 & 65.7 & 77.2 & 64.1 & 65.7 & 78.8 & 70.8 & 69.3 & 84.1\\
VIEScore & GPT-4o & 65.3 & 66.5 & 80.2  & 65.8 & 66.2 & 79.1  & 67.8 & 66.2 & 81.7 & 69.2 & 68.6 & 82.9 \\
\hline
\multicolumn{14}{c}{\textit{\textbf{Training-based Methods}}} \\
\hline
FGA-BLIP2 & BLIP2 & 62.1 & 64.6 & 76.8  & 56.6 & 57.9 & 71.4 & 63.2 & 65.1 & 77.7 & 65.3 & 66.9 & 79.0\\
\cline{2-14}
\multirow{3}{*}{SFT} & Qwen3-VL-8B-Instruct  & 58.6 & 57.1 & 72.2  & 63.4 & 63.9 & 76.7 & 65.2 & 66.7 & 79.3 & 67.1 & 65.7 & 80.3\\
& InternVL3-8B-hf  & 57.4 & 56.8 & 72.5  & 60.2 & 61.4 & 75.8 & 65.9 & 65.2 & 78.2 & 67.8 & 68.1 & 78.1\\
& LLaVA-v1.6-7B-hf & 54.7 & 55.2 & 73.1  & 55.7 & 57.4 & 70.9 & 61.7 & 62.5 & 74.3 & 62.9 & 63.5 & 76.5\\
\hline
\multicolumn{14}{c}{\textit{\textbf{\textsc{Revealer}}} \textit{\textbf{(Ours)}}} \\
\hline
& DINO + Gemini 3 Pro & 69.7 & 68.0 & 83.4  & 67.5 & 67.7 & 83.3 & 69.8 & 68.6 & 82.2 & 72.7 & \underline{74.0} & 84.5\\
\rowcolor{rowblue} \cellcolor{white} & \multicolumn{1}{l}{\scriptsize\textbf{\textit{\color{gaincolor}\hspace{1em} vs. Gemini 3 Pro}}} & 
\scriptsize\textbf{\color{gaincolor}(+1.6)} & \scriptsize\textbf{\color{gaincolor}(+2.2)} & \scriptsize\textbf{\color{gaincolor}(+2.1)} & 
\scriptsize\textbf{\color{gaincolor}(+1.4)} & \scriptsize\textbf{\color{gaincolor}(+2.3)} & \scriptsize\textbf{\color{gaincolor}(+2.5)} & 
\scriptsize\textbf{\color{gaincolor}(+1.3)} & \scriptsize\textbf{\color{gaincolor}(+1.4)} & \scriptsize\textbf{\color{gaincolor}(+1.2)} & 
\scriptsize\textbf{\color{gaincolor}(+1.4)} & \scriptsize\textbf{\color{gaincolor}(+1.7)} & \scriptsize\textbf{\color{gaincolor}(+0.6)} \\
\cdashline{2-14}
\cellcolor{white} & InternVL3-2B-hf & 64.7 & 65.3 & 77.8  & 63.5 & 64.2 & 78.6 & 64.7 & 63.4 & 77.9 & 66.8 & 66.3 & 79.4\\
\cellcolor{white} & InternVL3-8B-hf  & \underline{70.4} & \underline{69.5} & \underline{83.9}  & \underline{70.8} & \underline{69.2} & \underline{84.7} & \underline{70.6} & \underline{69.3} & \underline{83.4} & \underline{73.4} & 71.9 & \underline{85.0}\\
\rowcolor{rowblue} \cellcolor{white} & \multicolumn{1}{l}{\scriptsize\textbf{\textit{\color{gaincolor}\hspace{1em} vs. SFT}}} & 
\scriptsize\textbf{\color{gaincolor}(+13.0)} & \scriptsize\textbf{\color{gaincolor}(+12.7)} & \scriptsize\textbf{\color{gaincolor}(+11.4)} & 
\scriptsize\textbf{\color{gaincolor}(+10.6)} & \scriptsize\textbf{\color{gaincolor}(+7.8)} & \scriptsize\textbf{\color{gaincolor}(+8.9)} & 
\scriptsize\textbf{\color{gaincolor}(+4.7)} & \scriptsize\textbf{\color{gaincolor}(+4.1)} & \scriptsize\textbf{\color{gaincolor}(+5.2)} & 
\scriptsize\textbf{\color{gaincolor}(+5.6)} & \scriptsize\textbf{\color{gaincolor}(+3.8)} & \scriptsize\textbf{\color{gaincolor}(+6.9)} \\
\cdashline{2-14} 
\cellcolor{white} & Qwen3-VL-4B-Instruct  & 66.1 & 65.8 & 80.3  & 68.9 & 67.2 & 81.4 & 66.7 & 67.4 & 80.8 & 69.6 & 68.9 & 82.6\\
\cellcolor{white} \multirow{-6}{*}{\textbf{\textsc{Revealer}}} & Qwen3-VL-8B-Instruct  & \textbf{72.3} & \textbf{71.8} & \textbf{85.3} & \textbf{73.3} & \textbf{72.5} & \textbf{86.1} & \textbf{72.7} & \textbf{71.4} & \textbf{85.6} & \textbf{74.9} & \textbf{75.6} & \textbf{86.8}\\
\rowcolor{rowblue} \cellcolor{white} & \multicolumn{1}{l}{\scriptsize\textbf{\textit{\color{gaincolor}\hspace{1em} vs. SFT}}} & 
\scriptsize\textbf{\color{gaincolor}(+13.7)} & \scriptsize\textbf{\color{gaincolor}(+14.7)} & \scriptsize\textbf{\color{gaincolor}(+13.1)} & 
\scriptsize\textbf{\color{gaincolor}(+9.9)} & \scriptsize\textbf{\color{gaincolor}(+8.6)} & \scriptsize\textbf{\color{gaincolor}(+9.4)} & 
\scriptsize\textbf{\color{gaincolor}(+7.5)} & \scriptsize\textbf{\color{gaincolor}(+4.7)} & \scriptsize\textbf{\color{gaincolor}(+6.3)} & 
\scriptsize\textbf{\color{gaincolor}(+7.8)} & \scriptsize\textbf{\color{gaincolor}(+9.9)} & \scriptsize\textbf{\color{gaincolor}(+6.5)} \\
\bottomrule
\end{tabular}
\caption{Main results on element-level text-to-image alignment evaluation. We compare \textsc{Revealer} against representative Prompting-based and Training-based baselines. \textbf{Bold} and \underline{underlined} denote the best and second-best results, respectively. The rows labeled \textit{\color{gaincolor} vs.Gemini 3 Pro/SFT} highlight the absolute performance gains achieved by our method, demonstrating consistent and statistically significant improvements ($p=0.016 < 0.05$) over standard paradigms.}
\label{tab: main_result}
\end{table*}

\subsection{Main results}
Based on the results presented in Table \ref{tab: main_result}, we make the following observations.

First, the zero-shot adaptation of \textbf{\textsc{Revealer}} (combining Grounding DINO with Gemini 3 Pro) demonstrates superior performance compared to existing prompting-based baselines. 
It achieves comprehensive improvements over the strong TIFA (Gemini 3 Pro) baseline, with gains of \textbf{+1.6\%} SRCC, \textbf{+2.2\%} PLCC, and \textbf{+2.1\%} ACC on EvalMuse-40K, validating the effectiveness of the structured visual reasoning format itself.
Second, integrating GRPO training into \textbf{\textsc{Revealer}} yields substantial improvements over Training-based Methods. Specifically, on EvalMuse-40K, our InternVL3-8B-hf and Qwen3-VL-8B-Instruct models outperform their respective SFT counterparts by \textbf{+13.0\%} and \textbf{+13.7\%} in SRCC, \textbf{+12.7\%} and \textbf{+14.7\%} in PLCC, and \textbf{+11.4\%} and \textbf{+13.1\%} in ACC.
This indicates that RL effectively aligns the model's reasoning process with human preference beyond simple imitation learning.
Third, our best-performing model, \textbf{\textsc{Revealer}} (Qwen3-VL-8B-Instruct), achieves state-of-the-art performance across all metrics on all benchmarks. 
Compared to the strongest external proprietary baseline (TIFA with Gemini 3 Pro), our method establishes a clear margin, surpassing it by approximately \textbf{4.2\%} SRCC, \textbf{6.0\%} PLCC, and \textbf{4.0\%} ACC on EvalMuse-40K, and by \textbf{7.2\%} SRCC, \textbf{7.1\%} PLCC, and \textbf{5.3\%} ACC on RichHF.
Finally, notably, our \textbf{\textsc{Revealer}} models trained solely on the EvalMuse-40K dataset maintain stable high performance when evaluated on unseen benchmarks (RichHF, MHaluBench, and GenAI-Bench), demonstrating strong generalization capabilities beyond the training distribution.

\begin{table}[ht]
\centering
\renewcommand{\arraystretch}{1.0} 
\setlength{\tabcolsep}{4pt} 
\resizebox{1.0\linewidth}{!}{%
    \begin{tabular}{lcccc} 
    \toprule
    \multirow{2}{*}{\textbf{Model}} & \multicolumn{2}{c}{\textbf{EvalMuse-40K}} & \multicolumn{2}{c}{\textbf{RichHF}} \\
    \cmidrule(lr){2-3} \cmidrule(lr){4-5} 
     & \textbf{SRCC} & \textbf{ACC} & \textbf{SRCC} & \textbf{ACC}  \\
    \midrule
    
    Qwen3-VL-8B-Instruct & 55.7 & 65.9 & 56.2 & 70.7 \\  
    \hspace{1em} + Cold Start & 58.1 & 71.2 & 62.6 & 75.8 \\
    \hspace{2em} + Reasoning & 60.4 & 77.9 & 70.8 & 80.7 \\
    \hspace{3em} + Grouding & 59.8 & 72.1 & 67.5 & 78.2 \\
    \hspace{4em} \textbf{+ GRPO (\textbf{\textsc{Revealer}})} & \textbf{72.3} & \textbf{85.3} & \textbf{73.3} & \textbf{86.1} \\
    
    \midrule 
    
    \textbf{\textsc{Revealer}} & \textbf{72.3} & \textbf{85.3} & \textbf{73.3} & \textbf{86.1} \\
    \hspace{1em} w/o Visual Reasoning & 70.1 & 80.1 & 71.2 & 79.6 \\ 
    \hspace{1em} w/o Challenging Sample & 71.5 & 82.0 & 72.8 & 84.1 \\
    
    \bottomrule 
    \end{tabular}%
} 
\caption{Ablation studies on \textbf{EvalMuse-40K} and \textbf{RichHF} benchmarks (SRCC\% and Acc\%).} 
\label{tab: ablations}
\end{table}

\subsection{Ablations}
We conduct ablation studies to assess the contribution of each component in our framework. \textbf{Qwen3-VL-8B-Instruct} serves as the base model. ``+ Cold Start'' refers to supervised fine-tuning with formatted alignment data. ``+ Reasoning'' adds natural language explanation generation (\texttt{<thinking>}). ``+ Grounding'' introduces bounding box prediction (\texttt{<box>}) to ground observations before reasoning. ``+ GRPO'' (\textsc{Revealer}) applies reinforcement learning to align the model with human preferences. The subtractive settings ``w/o Visual Reasoning'' and ``w/o Challenging Sample'' denote the removal of the grouding step and the Challenging-sample Selection strategy during GRPO training, respectively.

As shown in Table \ref{tab: ablations}, performance generally improves with added components. Cold start and structured reasoning yield steady gains. GRPO brings the most significant improvement, with +13.2\% and +7.9\% accuracy gains on EvalMuse-40K and RichHF, respectively. Interestingly, visual reasoning without GRPO hurts performance, likely due to incorrect visual grounding leading to flawed reasoning. This is validated by the ``w/o Visual Reasoning'' setting, which results in drops of 5.2\% and 6.5\% on the two benchmarks. Finally, disabling challenging sampling leads to performance drops of 3.3\% and 2.0\%, confirming its positive effect on training quality.

\begin{figure}[t]
  \includegraphics[width=0.45\textwidth]{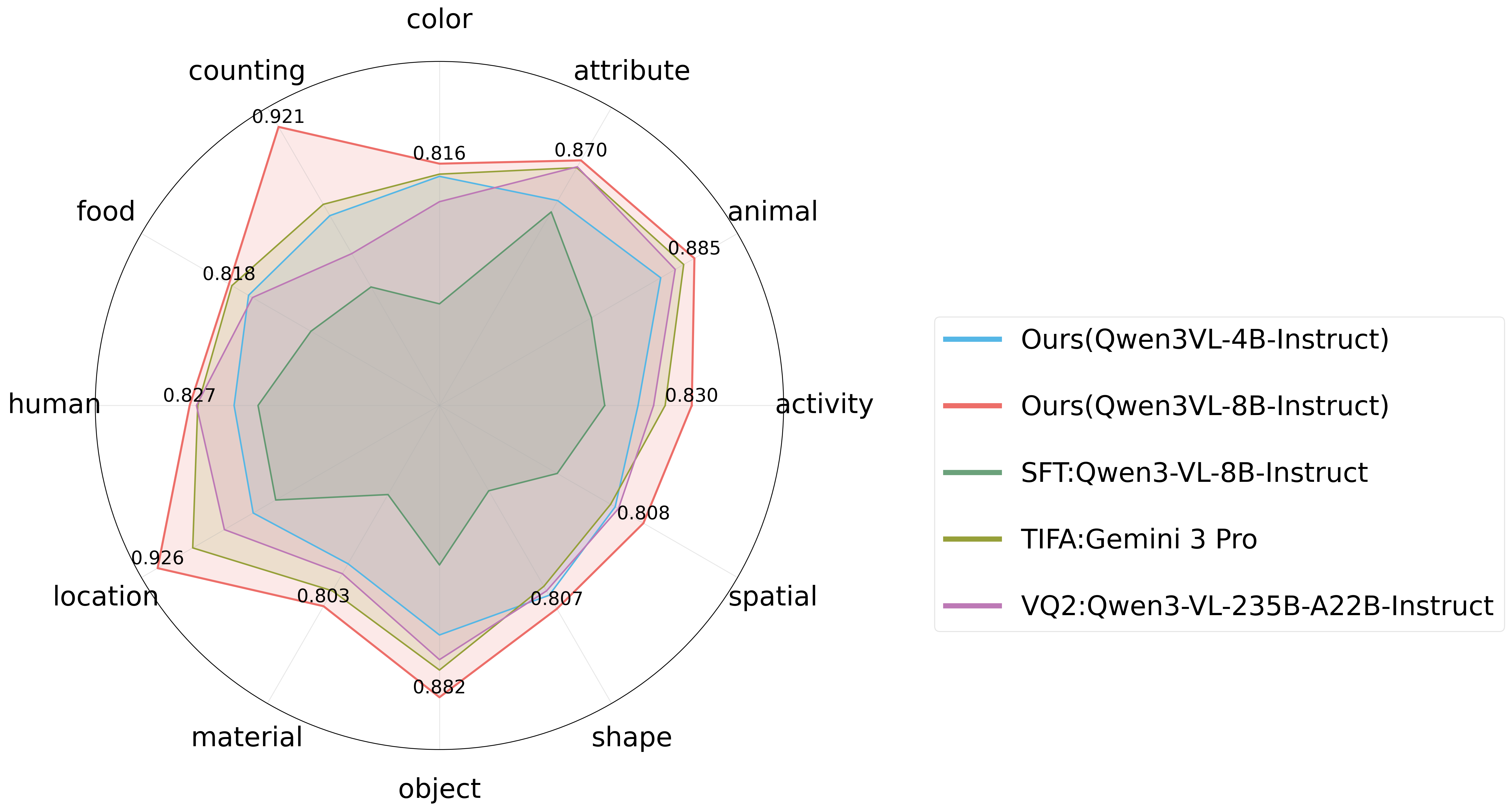}
  \caption{Accuracy across different element categories on the EvalMuse-40K benchmark.}
  \label{fig:acc}
\end{figure}

\subsection{Analyses}

\begin{table}[t]
    \centering
    \renewcommand{\arraystretch}{1.2}
    \resizebox{1.0\linewidth}{!}{
    \begin{tabular}{lcccccc}
        \toprule
        & & \multicolumn{2}{c}{\textbf{Empty Box Rate (\%)}} & \multicolumn{2}{c}{\textbf{Alignment Accuracy (\%)}} \\
        \cmidrule(lr){3-4} \cmidrule(lr){5-6}
        \textbf{Method} & \textbf{Threshold} & \textbf{Group A} & \textbf{Group B} & \textbf{Group A} & \textbf{Group B} \\
        & ($\gamma$) & (Concrete) & (Abstract) & (Concrete) & (Abstract) \\
        \midrule
        Baseline (Forced Grounding) & 0.35 & 2.1 & 12.4 & 86.3 & 80.5 \\
        \textbf{\textbf{\textsc{Revealer}} (strict Grounding)} & \textbf{0.55} & \textbf{4.5} & \textbf{53.0} & \textbf{87.2} & \textbf{84.7} \\
        \cdashline{1-7}
        \rowcolor{rowblue} \textit{\textbf{$\Delta$}} & - & \textbf{\color{gaincolor}+2.4} & \textbf{\color{gaincolor}+40.6} & \textbf{\color{gaincolor}+0.9} & \textbf{\color{gaincolor}+4.2} \\
        \bottomrule
    \end{tabular}
    }
    \caption{Impact of Strict Grounding Strategy ($\gamma=0.35 \rightarrow 0.55$). 
    Group A and B denote concrete and abstract elements, respectively.}
    \label{tab:abstract_analysis}
\end{table}

\noindent\textbf{Performance Across Different Element Categories.}
We evaluate alignment performance across different categories in EvalMuse-40K. As shown in Figure \ref{fig:acc}, our model (Qwen3-VL-8B-instruct) achieves superior performance, particularly in concrete categories like \textit{counting} and \textit{location}, validating the effectiveness of the structured \textbf{grounding–reasoning–conclusion} paradigm.

\noindent\textbf{Effect of Strict Grounding Strategy.} 
\label{sec:strict strategy}
To further address the challenge of localizing abstract concepts, we propose a \textit{Strict Grounding Strategy} by elevating the confidence threshold of Grounding DINO ($\gamma=0.35 \to 0.55$). 
This encourages the output of empty box lists (\texttt{[]}) when visual evidence is ambiguous, which prevents grounding error propagation and encourages the model to switch to global reasoning for abstract concepts.
As detailed in Table~\ref{tab:abstract_analysis}, this strategy increases the Empty Box Rate for abstract elements (Group B) from $12.4\%$ to $53.0\%$ while preserving precision for concrete ones (Group A). This strict Grounding mechanism yields a substantial \textbf{+4.2\%} accuracy gain on abstract attributes.

\begin{figure}[t]
  \includegraphics[width=0.45\textwidth]{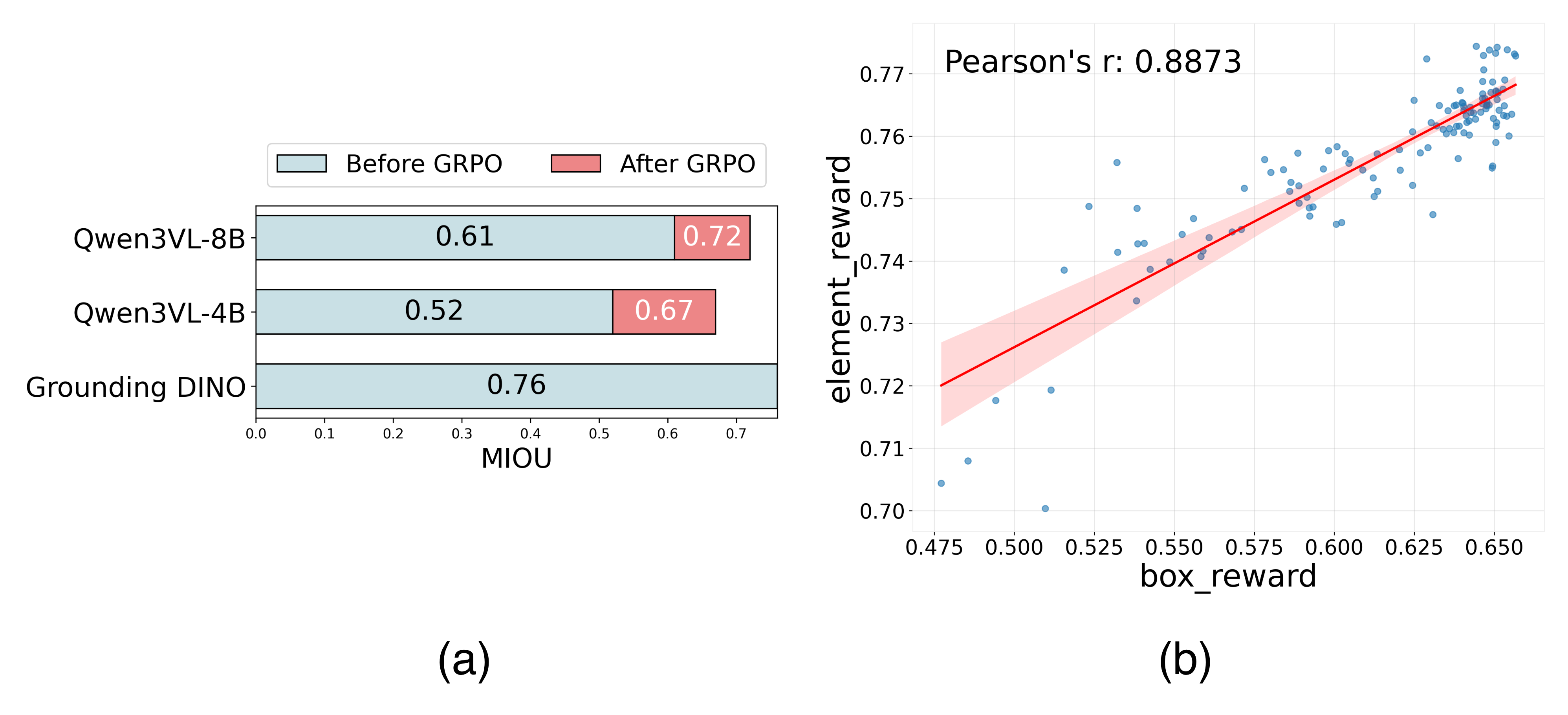}
  \caption{(a) Visual grounding capability before and after GRPO training. (b) Pearson correlation between box and element rewards.}
  \label{fig:analyes2}
\end{figure}

\noindent\textbf{Quality Validation of Visual Grounding Annotations.}
\label{sec:validation of Visual Grounding}
To validate the reliability of the bounding box annotations used in our automated data curation pipeline, and to provide a benchmark for evaluating visual grounding improvements, we constructed a high-quality, box-annotated evaluation set. 
Specifically, we constructed the evaluation set, denoted as $\mathcal{D}_{\text{BoxEval}}$, by randomly sampling a total of 2,000 image-prompt pairs from the EvalMuse-40K and RichHF benchmarks. We manually annotated the bounding boxes for the target elements within these pairs; notably, for abstract elements or elements absent from the image, we explicitly annotated the bounding box list as empty. To ensure precision, all annotations underwent a secondary review and correction process.
As illustrated in Figure~\ref{fig:analyes2}(a), we evaluated the performance of Grounding DINO on this human-verified set. 
The model achieved a mIoU of 0.76, indicating a high degree of overlap with human annotations. 
This result confirms the reliability of using Grounding DINO for large-scale training data synthesis.

\noindent\textbf{Visual Grounding Capability Before and After Training.}
We evaluate the visual grounding performance of our models on $\mathcal{D}_{\text{BoxEval}}$.
As shown in Figure~\ref{fig:analyes2}(a), we find that GRPO training significantly enhances localization capabilities, boosting mIoU by \textbf{+0.11} and \textbf{+0.15} for the 4B and 8B models, respectively.
Furthermore, we analyze the relationship between grounding precision and evaluation accuracy.
The results reveal a strong positive correlation (Pearson $r=0.8773$) between grounding accuracy and alignment scores (Figure~\ref{fig:analyes2}(b)), confirming that precise visual reasoning directly contributes to more accurate alignment evaluation.


\noindent\textbf{Visual Grounding Error Propagation Analysis.}To investigate error propagation from visual grounding to downstream reasoning, we conducted a detailed analysis using $\mathcal{D}_{\text{BoxEval}}$. Specifically, to ensure metric reliability, we manually verified the reasoning traces to identify hallucinations. As detailed in Table \ref{tab:error_propagation}, \textit{Misleading Grounding} (MIoU < 0.5) triggers severe error propagation, spiking the reasoning hallucination rate to \textbf{46.2\%} and drastically dropping alignment accuracy to \textbf{76.3\%}. This confirms that incorrect visual cues actively mislead the reasoning process. In contrast, our \textbf{Strict Grounding Strategy} acts as a safety mechanism by suppressing low-confidence predictions, effectively shifting high-risk samples to \textit{Global Reasoning}. This fallback mechanism significantly reduces hallucinations to \textbf{14.7\%} and recovers alignment accuracy to \textbf{81.2\%}, demonstrating that relying on global context is far superior to reasoning based on erroneous visual evidence.

\begin{table}[t]
\centering
\small
\renewcommand{\arraystretch}{1.3} 
\setlength{\tabcolsep}{4pt} 

\resizebox{1.0\linewidth}{!}{%
    \begin{tabular}{l l c c c}
    \toprule
    \multirow{2}{*}{\textbf{Grounding Status}} & \textbf{Condition} & \multirow{2}{*}{\textbf{Distribution}} & \textbf{Reasoning Hal.} & \textbf{Alignment} \\
    & \textit{(Filter Criteria)} &  & \textbf{Rate$\downarrow$ } & \textbf{Acc$\uparrow$} \\
    \midrule
    Accurate Grounding & MIoU $\ge$ 0.5 & 80.5\% & \textbf{8.4\%} & \textbf{89.6\%} \\
    Misleading Grounding & MIoU < 0.5 & 8.1\% & 46.2\% & 76.3\% \\
    \rowcolor{rowblue} \textbf{Strict Grounding Strategy } & Empty Box (\texttt{[]})& 12.4\% & 14.7\% & \underline{81.2\%} \\
    
    \bottomrule
    \end{tabular}%
}
\caption{Visual Grounding Error Propagation Analysis (Qwen3-VL-8B).} 
\label{tab:error_propagation}
\end{table}
\section{Conclusion}
We introduced \textbf{\textsc{Revealer}}, a reinforcement-guided visual reasoning framework for element-level text-to-image alignment evaluation. By enforcing a structured ``grounding–reasoning–conclusion'' paradigm and optimizing via GRPO, our approach effectively bridges the gap between visual localization and semantic judgment. Experiments across four benchmarks show that \textsc{Revealer} achieves state-of-the-art performance, surpassing proprietary models like Gemini 3 Pro. 


\section*{Limitations}
Despite the superior performance of \textsc{Revealer}, several limitations remain. First, the explicit box-based grounding paradigm is optimized for concrete semantic elements and may be less naturally suited for evaluating holistic qualities, such as artistic style, complex lighting atmospheres, or emotional tone, where discrete localization is ambiguous. Furthermore, our current work focuses exclusively on static image-text alignment; consequently, the applicability of our framework to text-to-video alignment evaluation is limited, as it does not account for temporal dynamics or motion consistency. Future work will aim to extend the visual reasoning framework to address these non-localized and temporal challenges.
\bibliography{custom}
\clearpage

\appendix
\section{Dataset Details}
\subsection{Quality Assurance for Visual Reasoning Trajectory}
\label{app:quality_assurance}
To ensure the high quality of the generated reasoning rationales $r_i$, we employ a two-stage quality assurance strategy to strictly filter out low-quality samples. 
First, in the \textit{self-correction stage}, if the predicted label $\hat{a}_i$ is inconsistent with the ground-truth label $a_i$, we re-prompt the model to generate a revised explanation and prediction. Data points that fail to reach consistency after three attempts are discarded, and for the retained samples, we adopt the human-annotated label $a_i$ from EvalMuse-40K as the final ground truth. 
Second, in the \textit{logical verification stage}, we employ Gemini 3 pro to further guarantee logical coherence by verifying the consistency between the generated $r_i$ and the label $a_i$. Specifically, the model assesses whether $r_i$ logically supports $a_i$, and any data points exhibiting logical inconsistencies are strictly filtered out.

\section{Training Details}
\label{sec:training_objectives}

\begin{figure}[h]
  \centering
  \includegraphics[width=0.8\linewidth]{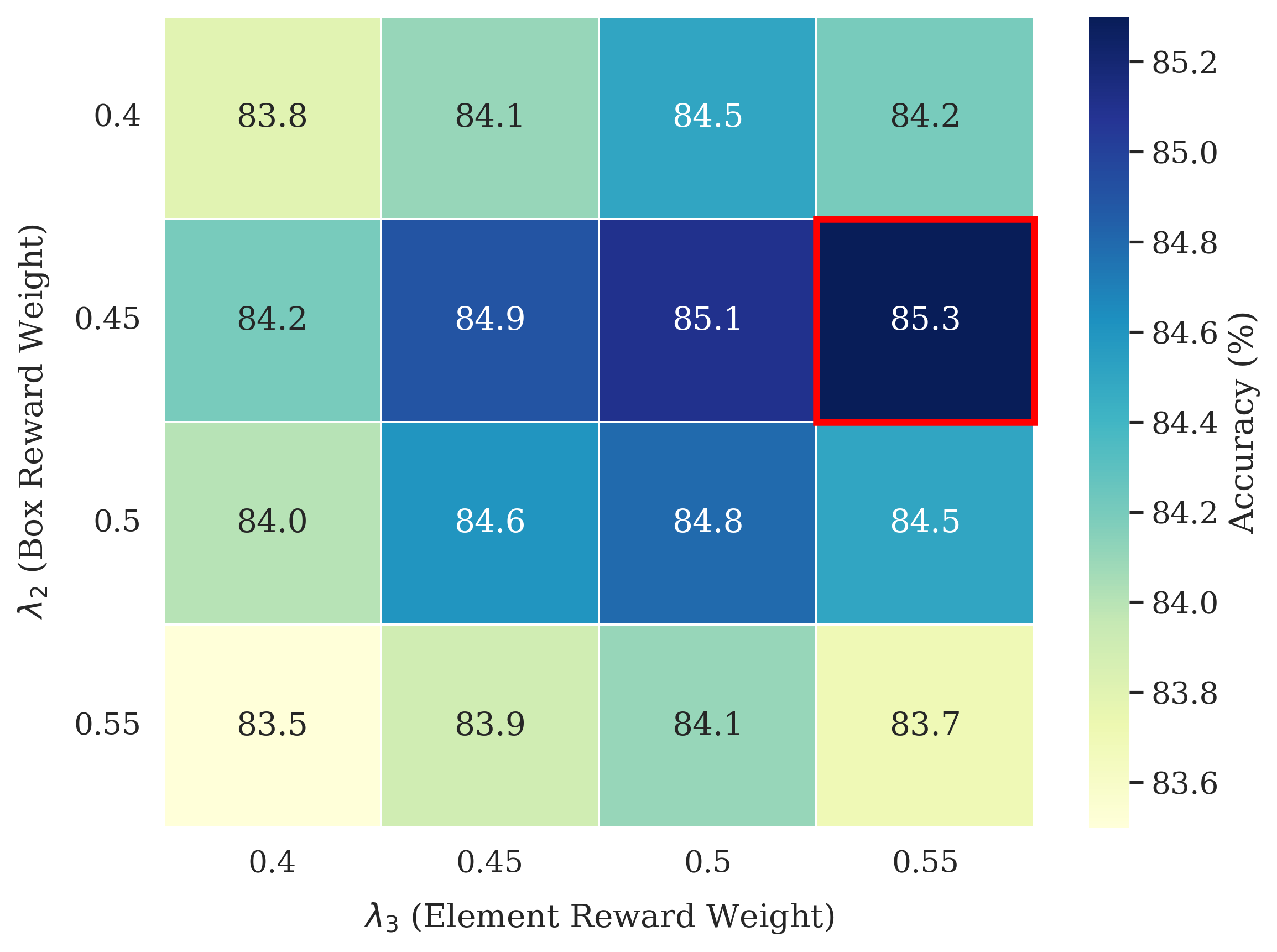}
  \caption{Grid search results for reward weights $\lambda_2$ and $\lambda_3$ with fixed $\lambda_1 = 0.1$. The heatmap shows validation accuracy on EvalMuse-40K. The red box indicates the optimal configuration ($\lambda_2=0.45, \lambda_3=0.55$).}
  \label{fig:hyperparam_heatmap}
\end{figure}

\paragraph{Hyperparameter Sensitivity and Configuration}
To balance the multi-objective nature of our reward function, we conducted a grid search to determine the optimal scalar coefficients $\lambda_1$, $\lambda_2$, and $\lambda_3$. 
We observed that the model quickly learns to adhere to the structural format; therefore, we fixed the format reward weight at a low value of $\lambda_1 = 0.1$ to prevent it from dominating the optimization landscape. 
We then performed a grid search for the visual grounding weight ($\lambda_2$) and element alignment weight ($\lambda_3$) over the range $\{0.4, 0.45, 0.5, 0.55\}$. We evaluated the model's performance on a hold-out validation set from EvalMuse-40K. As illustrated in Figure~\ref{fig:hyperparam_heatmap}, the results indicate a performance peak where slightly higher emphasis is placed on the final element alignment score. The optimal configuration was identified as $\lambda_1 = 0.1$, $\lambda_2 = 0.45$, and $\lambda_3 = 0.55$. This setting ensures that while visual grounding provides necessary evidence, the ultimate fidelity of the alignment judgment remains the primary optimization target.

\paragraph{SFT Objective.}
In the cold-start stage, we fine-tune the model to generate the structured reasoning trajectory. Let $q$ denote the concatenation of the input inputs $(\mathcal{I}, \mathcal{P}, \{ e_i \}_{i=1}^N)$, and $g$ denote the target output sequence formed by concatenating $\{\langle b_i, r_i, a_i \rangle\}_{i=1}^N$. The training objective is to minimize the negative log-likelihood:
\begin{equation}
\mathcal{L}_{\text{cold}} = - \mathbb{E}_{q \sim \mathcal{D}_{SFT}}\sum_{t=1}^{T} \log P_\theta(g_t \mid g_{<t}, q)
\end{equation}
where $g_t$ is the $t$-th token in the output sequence and $\theta$ denotes the model parameters.

\paragraph{GRPO Optimization.}
Given the defined total reward $r(\tau)$, we optimize the policy model using GRPO, a lightweight and stable variant of Proximal Policy Optimization (PPO). 
Specifically, for each $(\mathcal{I}, \mathcal{P}, \{\langle e_i, b_i, r_i, a_i \rangle\}_\text{i=1}^N)$ in $\mathcal{D}_\text{Challenging-Sample}$, a reasoning trajectory sequence $\tau$ generated by the policy model, the rule-based reward function $r(\cdot)$ computes its reward as $r(\tau)$. 
GRPO normalizes this scalar into an advantage $A_t = \frac{r(\tau) - \mu}{\sigma}$ for each decoding step $t \in \{1, \ldots, T\}$, where $\mu$ and $\sigma$ are the batch-wise mean and standard deviation of rewards.
GRPO samples a group of generated output set $\{o_1, o_2, \cdots, o_G\}$ for each $q$ from the policy model $\pi_{\theta_{\text{old}}}$ and let the policy ratio at step $t$ be $\rho_t = \frac{\pi_\theta(o_t \mid o_{i,<t}, q)}{\pi_{\theta_{\text{old}}}(o_t \mid o_{i,<t}, q)}$, where $o_i$ represents the outputs sampled from the policy model. 
The trained policy $\pi_{\theta}$ is then updated by maximizing the following objective:

\begin{equation}
\footnotesize
\small
\begin{split}
\mathcal{J}_{GRPO}(\theta) &= \mathbb{E}{[q \sim \mathcal{D}_\text{Challenging-Sample}, \{o_i\}_{i=1}^G \sim \pi_{\theta_{old}}(O|q)]} \\
&\quad \frac{1}{G}\sum_{i=1}^G\frac{1}{|o_i|} \sum_{t=1}^{|o_i|} \left\{ \min \left[ \rho_t A_t, \text{clip}(\cdot) \cdot A_t \right] \right. \\
&\quad \left. - \beta \mathbb{D}_{\text{KL}}\left[\pi_{\theta} || \pi_{\text{ref}}\right]\right\} ,
\end{split}
\label{eq:GRPO-obj}
\end{equation}
Here, $\pi_{\text{ref}}$ denotes the frozen reference policy obtained from the SFT stage.
$\text{clip}(\cdot)$ refers to applying a clipping function to $\rho_t$ that bounds it within $[1 - \epsilon, 1 + \epsilon]$, where $\epsilon$ is hyperparameter. 
This clip function helps prevent excessively large policy updates. 
Unlike the KL penalty term used in \citep{ouyang2022training}, we estimate the KL divergence with the unbiased estimator \citep{kl_approx}, which is guaranteed to be positive.
We set  $\epsilon = 0.2$ and $\beta = 1e - 2$  during training.
The hyperparameter $\beta$ controls the KL divergence penalty, which encourages the new policy to stay close to the reference policy, thereby stabilizing training.

\section{Evaluation Details}
\subsection{Benchmark Details}
\label{sec: benchmark}
We evaluate alignment accuracy across four fine-grained benchmarks: EvalMuse-40K, RichHF, MHaluBench, and GenAI-Bench. (1) \textbf{EvalMuse-40K} provides element-level alignment annotations across categories such as object, attribute, and location. Each element is labeled as aligned $(1)$ or unaligned $(0)$ by multiple annotators, and final labels are averaged; elements with scores $\geq 0.5$ are considered aligned. (2) \textbf{RichHF}~\cite{richhf} offers keyword-level annotations over diverse prompt styles. We evaluate on the annotated subset using accuracy. (3) \textbf{MHaluBench}~\cite{unifiedhallucination} provides claim-level annotations. To enable fine-grained evaluation, we extract elements via GPT-4~\cite{gpt4}, generate binary questions, and collect human annotations following the EvalMuse-40K protocol. (4) \textbf{GenAI-Bench}~\cite{genaibench} targets complex compositional prompts. As it lacks element-level labels, we apply the same procedure as in MHaluBench.

\subsection{Adaptation of Benchmarks for Fine-Grained Evaluation}
To support element-level multimodal hallucination detection, we adapted two existing benchmarks---\textbf{MHaluBench} and \textbf{GenAI-Bench}---by applying a unified annotation protocol inspired by EvalMuse-40K~\cite{evalmuse}. While MHaluBench (specifically its text-to-image subset) and GenAI-Bench provide diverse prompting schemes, they originally lack granular semantic annotations. To address this, we decompose each natural language prompt into discrete semantic elements using GPT-4, categorizing them according to the TIFA taxonomy (e.g., object, attribute, spatial). For each element, we generate a corresponding binary verification question (e.g., ``Is there a red car in the image?'') to assess visual fidelity. These element-question pairs undergo rigorous human verification to determine semantic alignment (labeled as 1 for aligned, 0 for misaligned), thereby enabling consistent, interpretable, and fine-grained evaluation across both compositional and general scenarios.

\subsection{Adaptation of Zero-Shot Baselines for Element-Level Evaluation}

To ensure a rigorous comparison, we adapt representative zero-shot methods—TIFA, VQ$^2$, VQAScore, and VIEScore—to our fine-grained evaluation task through a unified pipeline. For each baseline, we first employ a large language model (GPT-4) to decompose the input prompt into discrete, visually verifiable semantic units according to the TIFA taxonomy, such as objects, attributes, and spatial. These units are subsequently converted into method-specific query formats, ranging from binary VQA questions to structured semantic triples. Finally, pre-trained multimodal models are utilized to verify the visual grounding of each query against the generated image. This process standardizes the output into binary alignment labels for individual semantic elements, facilitating a consistent and interpretable performance assessment across all methods.

\section{Additional Analyses}
\label{sec:appendix}

\setlength{\tabcolsep}{0.35mm}
\begin{table}[t]
    \small
    \centering
    { 
     \begin{tabular}{c w{c}{0.8cm} w{c}{0.8cm}  w{c}{0.8cm} w{c}{0.8cm}  w{c}{0.8cm} w{c}{0.8cm}}
        \toprule
        \multirow{2}{*}{\textbf{Method}} & \multicolumn{2}{c}{\textbf{RichHF}} & \multicolumn{2}{c}{\textbf{MHaluBench}} &
        \multicolumn{2}{c}{\textbf{GenAI-Bench}} \\ 
        \cmidrule(lr){2-3} \cmidrule(lr){4-5} \cmidrule(lr){6-7}
        & srcc$\uparrow$ & time$\downarrow$ & srcc$\uparrow$ & time$\downarrow$ & srcc$\uparrow$ & time$\downarrow$ \\
        \midrule
        Chain-of-Focus & 65.1 & 5.9  & 67.3 & 7.3 & 69.1 & 6.6\\
        ViLaSR & 64.2 & 6.5 & 65.4 & 5.7 & 68.9 & 6.2\\
        Vision-R1 & 64.7 & 5.9  & 66.2 & 6.8 & 68.0  & 4.8\\
        Q-Insight & \underline{67.4} & \underline{4.5}  & \underline{67.9} & \underline{4.1} & \underline{70.3}  & \underline{4.4}\\
        \midrule
        \rowcolor{rowblue} \textbf{\textsc{Revealer}} & \textbf{70.8} & \textbf{1.3} & \textbf{70.6} & \textbf{1.6} & \textbf{74.4} & \textbf{1.2} \\
        \bottomrule
    \end{tabular}
    }
    \caption{Comparison with RL-based visual reasoning methods. \textbf{Time} denotes the average inference latency per sample measured on a single A800 GPU.}
    \label{tab:rl_comparison}
\end{table}

\subsection{Comparison with RL-based Visual Reasoning Methods.}
We compare \textsc{Revealer} against representative RL-based MLLMs, including Chain-of-Focus~\cite{cof}, ViLaSR~\cite{vilasr}, Vision-R1~\cite{visionr1}, and Q-Insight~\cite{qinsight}. As shown in Table~\ref{tab:rl_comparison}, our method establishes a superior trade-off between alignment accuracy and computational efficiency.
\textbf{Accuracy.} Existing iterative methods (Chain-of-Focus, ViLaSR) suffer from error propagation during multi-turn interactions, while Q-Insight focuses more on the global image quality score. By anchoring reasoning to specific elements via a structured paradigm, \textsc{Revealer} mitigates these issues, surpassing the strongest baseline (Q-Insight) by significant margins of \textbf{+3.4\%} and \textbf{+4.1\%} SRCC on RichHF and GenAI-Bench, respectively.
\textbf{Efficiency.} Unlike baselines that require multiple forward passes for visual resampling or unconstrained reasoning generation, \textsc{Revealer} integrates localization and reasoning into a single cohesive pass. This streamlined architecture reduces inference time to \textbf{1.2s--1.6s} per sample, representing a significant efficiency gain over preceding RL-based methods.

\subsection{Impact of Continuous vs. Binary Rewards on GRPO Training.}
\begin{figure}[t]
  \includegraphics[width=0.47\textwidth]{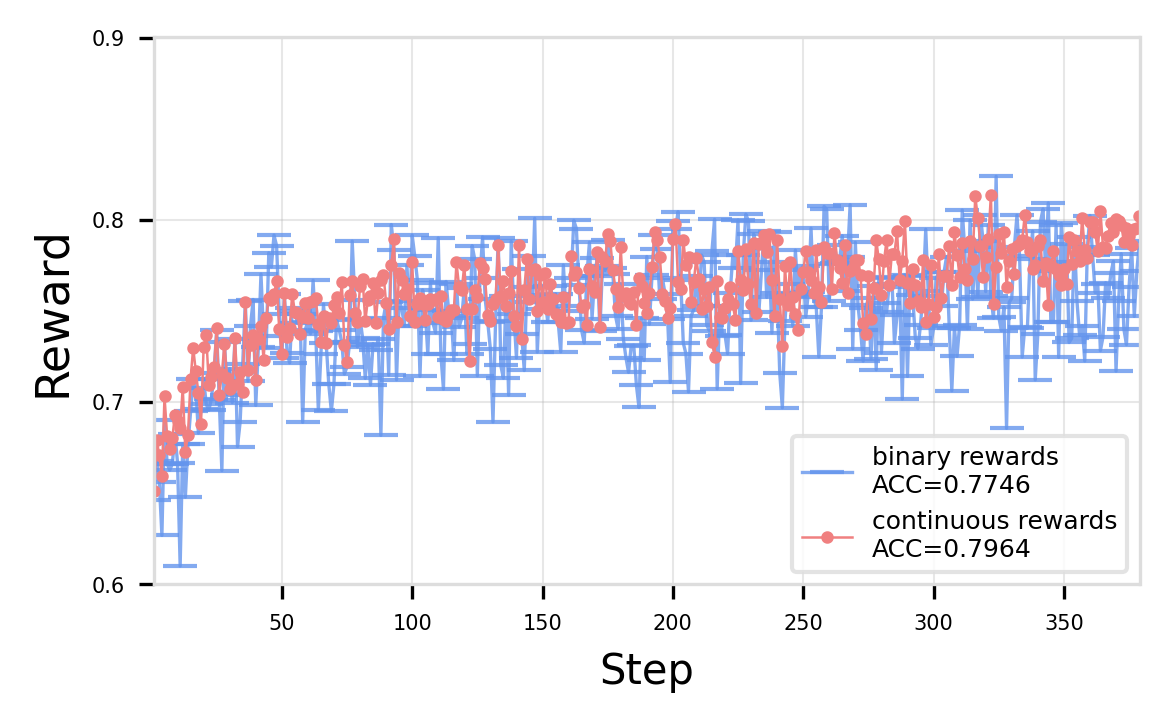}
  \caption{Comparison of training stability and final accuracy between binary and continuous reward designs. Continuous rewards result in more stable training and better alignment performance on EvalMuse-40K, achieving an ACC of 0.7964 compared to 0.7746 with binary rewards.}
  \label{fig:element_reward}
\end{figure}

To investigate the impact of reward formulation on GRPO training, we compare two designs of element-level reward for Qwen2.5-VL-3B-Instruct. The first is a binary reward, where each element is assigned 1 if the alignment prediction is correct and 0 otherwise. The second is a continuous reward, calculated as the absolute difference between the model’s predicted alignment score, bounded within $[0,1]$, and the corresponding ground-truth label. As shown in Figure~\ref{fig:element_reward}, training with continuous rewards yields a more stable optimization process and outperforms binary rewards by 2.2\% in accuracy on the EvalMuse-40K benchmark. We attribute this improvement to the finer-grained feedback provided by continuous rewards. In particular, continuous rewards lead to smoother reward landscapes and more reliable policy updates, especially during early training when binary signals are often sparse or uninformative. 

\subsection{Statistical Significance Analysis}
To rigorously validate that the performance gains of our proposed method over strong baselines (DINO, Gemini 3 Pro) stem from the effective reinforcement-guided reasoning framework rather than random variance, we conducted a statistical significance test using a stratified bucketing approach. Specifically, we randomly partitioned the EvalMuse-40K test set into $K=10$ disjoint folds and computed the element-level alignment accuracy for both our GRPO-optimized model (Qwen3-VL-8B-Instruct) and the Gemini 3 Pro baseline on each fold. We then performed a one-sided Wilcoxon Signed-Rank Test on the resulting paired accuracy distributions to assess the consistency of the improvement. The analysis yielded a $p$-value of \textbf{0.016} ($p < 0.05$). This result statistically rejects the null hypothesis, confirming that our method's improvement is statistically significant and robust across different data distributions.

\section{Prompt Templates}
\label{app: prompt_templates}

\begin{figure*}[t!]
    \centering
    \begin{tcolorbox}[
        title=\textbf{Prompt for Visual Reasoning Trajectory Generation},
        colback=gray!5,
        colframe=black!70,
        boxrule=0.8pt,
        width=\textwidth,
        arc=2mm,
        left=2pt, right=2pt
]
\begin{lstlisting}[basicstyle=\small\ttfamily, columns=fullflexible, breaklines=true, breakatwhitespace=false, keepspaces=true, frame=none, aboveskip=0pt, belowskip=0pt, mathescape=true]
System Instruction:
You are an expert evaluator for text-to-image alignment. Your task is to perform visual reasoning to determine if a specific element ($e_i$) from the input prompt ($\mathcal{P}$) is accurately represented in the generated image ($\mathcal{I}$). You are provided with bounding boxes ($b_i$) detected by a grounding model.

Input Data:
- Full Prompt ($\mathcal{P}$): {full_prompt}
- Target Element ($e_i$): {element}
- Bounding Boxes ($b_i$): {box_list} (Format: [[x1, y1, x2, y2]...])

Reasoning Rules:
1. Localized Reasoning (If $b_i$ is NOT empty):
   - Focus strict attention on the visual content within the provided coordinates.
   - Verify if the visual element inside the boxes match the description of $e_i$.
   - Ignore background details outside the boxes unless they directly affect the element's state.

2. Global Reasoning (If $b_i$ is empty []):
   - Switch to Global Context Analysis. The grounding model failed to localize the element.
   - Scenario A (Concrete Object): If $e_i$ is a tangible object (e.g., "cat", "car"), its absence usually implies misalignment. Verify if it is truly missing.
   - Scenario B (Abstract Attribute/Style): If $e_i$ is global (e.g., "foggy", "oil painting", "lighting"), evaluate the entire image atmosphere. Empty boxes are expected here.

Output Format:
Return a JSON object containing:
- "reasoning" ($r_i$): A step-by-step rationale based on the rules above. 
- "label" ($\hat{a}_i$): 1 for Aligned, 0 for Misaligned.

Response:
\end{lstlisting}
    \end{tcolorbox}
    \caption{The system prompt template used for visual reasoning trajectory curation. The prompt explicitly instructs the model to handle both grounded (localized) and ungrounded (global) scenarios.}
    \label{fig:prompt_template}
\end{figure*}

\begin{figure*}[t!]
    \centering
    \begin{tcolorbox}[
        title=\textbf{Prompt for Visual Reasoning Self-Correction},
        colback=gray!5,
        colframe=black!70,
        boxrule=0.8pt,
        width=\textwidth,
        arc=2mm,
        left=2pt, right=2pt
    ]
\begin{lstlisting}[basicstyle=\small\ttfamily, columns=fullflexible, breaklines=true, breakatwhitespace=false, keepspaces=true, frame=none, aboveskip=0pt, belowskip=0pt, mathescape=true]
System Instruction:
You are an expert evaluator for text-to-image alignment. You are provided with a Reference Alignment Label ($a_i$) derived from human annotation for a specific element ($e_i$). Your task is to re-examine the image and bounding boxes ($b_i$) to construct a visual reasoning path that logically supports this reference label.

Input Data:
- Full Prompt ($\mathcal{P}$): {full_prompt}
- Target Element ($e_i$): {element}
- Bounding Boxes ($b_i$): {box_list}
- Reference Label ($a_i$): {ground_truth_label} (1 = Aligned, 0 = Misaligned)

Reasoning Rules:
1. Localized Reasoning (If $b_i$ is NOT empty):
   - Focus strict attention on the visual content within the provided coordinates.
   - Verify if the visual element inside the boxes match the description of $e_i$.
   - Ignore background details outside the boxes unless they directly affect the element's state.

2. Global Reasoning (If $b_i$ is empty []):
   - Switch to Global Context Analysis. The grounding model failed to localize the element.
   - Scenario A (Concrete Object): If $e_i$ is a tangible object (e.g., "cat", "car"), its absence usually implies misalignment. Verify if it is truly missing.
   - Scenario B (Abstract Attribute/Style): If $e_i$ is global (e.g., "foggy", "oil painting", "lighting"), evaluate the entire image atmosphere. Empty boxes are expected here.

Correction Rules:
1. Evidence Re-Discovery:
   - If Reference ($a_i$) is 1 (Aligned): Look closely at the region/image to identify the specific visual features (color, shape, count) that confirm the element's presence.
   - If Reference ($a_i$) is 0 (Misaligned): Look for the specific visual discrepancy (e.g., wrong color, missing object, distorted shape) that contradicts the prompt.

2. Strict Formatting Constraint (Crucial):
   - Your reasoning must be self-contained and based solely on visual observation.
   - DO NOT mention the "Reference Label," "Human Annotation," or "Ground Truth" in your reasoning text.
   - DO NOT write phrases like "As indicated by the reference..." or "Since the label is 1..."
   - Simply state the visual facts that lead to the conclusion.

Output Format:
Return a JSON object containing:
- "reasoning" ($r_i$): A factual visual analysis describing *why* the image conforms to the Reference Label.
- "label" ($\hat{a}_i$): The final label (should match $a_i$).

Response:
\end{lstlisting}
    \end{tcolorbox}
    \caption{The self-correction prompt template. When the initial prediction disagrees with the ground truth, the model is guided to re-evaluate the visual evidence to align with the human annotation ($a_i$) without explicitly referencing the hint in the rationale.}
    \label{fig:correction_prompt}
\end{figure*}

\begin{figure*}[t!]
    \centering
    \begin{tcolorbox}[
        title=\textbf{Prompt for Logical Consistency Verification},
        colback=gray!5,
        colframe=black!70,
        boxrule=0.8pt,
        width=\textwidth,
        arc=2mm,
        left=2pt, right=2pt
    ]
\begin{lstlisting}[basicstyle=\small\ttfamily, columns=fullflexible, breaklines=true, breakatwhitespace=false, keepspaces=true, frame=none, aboveskip=0pt, belowskip=0pt, mathescape=true]
System Instruction:
You are a Quality Assurance Auditor for an automated evaluation system. Your task is to verify the logical consistency between a generated reasoning rationale ($r_i$) and its assigned binary label ($a_i$) for a target element ($e_i$). You must detect contradictions between the textual explanation and the numerical score.

Input Data:
- Target Element ($e_i$): {element}
- Generated Reasoning ($r_i$): {reasoning_text}
- Assigned Label ($a_i$): {label} (1 = Aligned, 0 = Misaligned)

Verification Rules:
1. Logical Entailment Check:
   - Does the text in $r_i$ explicitly state that the element is correctly depicted or aligned? If yes, $a_i$ must be 1.
   - Does the text in $r_i$ describe missing objects, wrong attributes, or hallucinations? If yes, $a_i$ must be 0.

2. Identify Contradictions:
   - Flag as "Inconsistent" if $r_i$ describes a failure (e.g., "The car is blue instead of red") but $a_i$ is 1.
   - Flag as "Inconsistent" if $r_i$ describes a success (e.g., "The car is correctly rendered in red") but $a_i$ is 0.

Output Format:
Return a JSON object containing:
- "is_consistent": boolean (true/false)
- "analysis": "Brief explanation of the consistency check."

Response:
\end{lstlisting}
    \end{tcolorbox}
    \caption{The logical verification prompt used by Gemini 3 Pro. This step filters out low-quality samples where the generated reasoning text ($r_i$) logically contradicts the final classification label ($a_i$).}
    \label{fig:verification_prompt}
\end{figure*}

\begin{figure*}[t!]
    \centering
    \begin{tcolorbox}[
        title=\textbf{Prompt for End-to-End Element Alignment Inference},
        colback=gray!5,
        colframe=black!70,
        boxrule=0.8pt,
        width=\textwidth,
        arc=2mm,
        left=2pt, right=2pt
    ]
\begin{lstlisting}[basicstyle=\small\ttfamily, columns=fullflexible, breaklines=true, breakatwhitespace=false, keepspaces=true, frame=none, aboveskip=0pt, belowskip=0pt, mathescape=true]
<image>
System Instruction:
You are an expert in fine-grained text-to-image alignment evaluation. Your task is to perform Element-level Hallucination Detection on the provided image based on the input prompt.

Input Data:
- Prompt ($\mathcal{P}$): {original_prompt}
- Target Elements ($E$): {element_keys_str}

Evaluation Protocol:
For each element in the target list, perform the following steps sequentially:
1. Localization (<box>):
   - Identify the element's location in the image.
   - Output bounding boxes in the format [[x1, y1, x2, y2]...] .
   - If the element is missing or abstract (unable to be grounded), output an empty list [].

2. Visual Reasoning (<thinking>):
   - Analyze whether the visual depiction matches the textual description (appearance, action, relation).
   - Explicitly state any discrepancies (e.g., "present but wrong color", "missing entirely").

3. Scoring (<score>):
   - Assign a fidelity score between 0.0 and 1.0.
   - 1.0 = Perfectly present and accurate.
   - 0.0 = Entirely missing or hallucinated.
   
Output Format:
Output a single Python dictionary string wrapped in <element> tags.
- Keys: Element names (Categories).
- Values: A concatenated string containing the tags <box>...</box><thinking>...</thinking><score>...</score>.

Example Output:
<element>
{
  "Eating (activity)": "<box>[[221, 162, 893, 675]]</box><thinking>The subject has food but is not performing the action of eating.</thinking><score>0.4</score>",
  "Puffin (animal)": "<box>[[1, 10, 486, 365]]</box><thinking>The puffin is rendered clearly but is in the wrong spatial location.</thinking><score>0.3</score>",
  "Pink tree (object)": "<box>[[122, 95, 900, 883]]</box><thinking>The tree matches the color and style description perfectly.</thinking><score>1.0</score>"
}
</element>

Constraint:
Do not include any conversational text outside the <element> tags. Ensure the JSON syntax is valid.

Response:
\end{lstlisting}
    \end{tcolorbox}
    \caption{The inference prompt used for evaluating text-to-image models. It enforces a strict "Grounding-Reasoning-Scoring" format output within a structured dictionary for automated parsing.}
    \label{fig:inference_prompt}
\end{figure*}

\end{document}